\documentclass[10pt,journal,compsoc]{IEEEtran}
\usepackage{amsmath,amsfonts,amssymb}
\usepackage{algorithmic}
\usepackage{algorithm}
\usepackage{array}
\usepackage[caption=false,font=normalsize,labelfont=sf,textfont=sf]{subfig}
\usepackage{textcomp}
\usepackage{stfloats}
\usepackage{url}
\usepackage{verbatim}
\usepackage{graphicx}
\usepackage{cite}

\usepackage{multirow}
\usepackage{booktabs}
\usepackage{color} 
\begin{document}
\title{PointCG: Self-supervised Point Cloud Learning via Joint Completion and Generation}

\author{Yun Liu, Peng Li, Xuefeng Yan, Liangliang Nan, Bing Wang, Honghua Chen, Lina Gong, Wei Zhao, and Mingqiang Wei,~\textit{Senior Member, IEEE}

\thanks{Yun Liu, Peng Li, Honghua Chen, Lina Gong, Wei Zhao and Mingqiang Wei are with the School of Computer Science and Technology, Nanjing University of Aeronautics and Astronautics, Nanjing, China, and also with the Shenzhen Institute of Research, Nanjing University of Aeronautics and Astronautics, Shenzhen, China (e-mail: yun.liu.lydia@gmail.com, pengl@nuaa.edu.cn, chenhonghuacn@gmail.com, gonglina@nuaa.edu.cn, weizhao0120@nuaa.edu.cn, mingqiang.wei@gmail.com). }

\thanks{Xuefeng Yan is with the School of Computer Science and Technology, Nanjing University of Aeronautics and Astronautics, Nanjing, China, and also with the Collaborative Innovation Center of Novel Software Technology and Industrialization, Nanjing, China (e-mail: fzjm\_313@nuaa.edu.cn).}

\thanks{Liangliang Nan is with Urban Data Science section, Delft University of Technology, Delft, Netherlands (e-mail: liangliang.nan@gmail.com). }

\thanks{Bing Wang is with the Department of Aeronautical and Aviation Engineering, The Hong Kong Polytechnic University, Hongkong, China (e-mail: bingwang@polyu.edu.hk). }

}
\markboth{Journal of \LaTeX\ Class Files,~Vol.~14, No.~8, August~2021}%
{Shell \MakeLowercase{\textit{et al.}}: A Sample Article Using IEEEtran.cls for IEEE Journals}


\IEEEtitleabstractindextext{%
\begin{abstract}

The core of self-supervised point cloud learning lies in setting up appropriate pretext tasks, to construct a pre-training framework that enables the encoder to perceive 3D objects effectively. 
In this paper, we integrate two prevalent methods, masked point modeling (MPM) and 3D-to-2D generation, as pretext tasks within a pre-training framework. 
We leverage the spatial awareness and precise supervision offered by these two methods to address their respective limitations: ambiguous supervision signals and insensitivity to geometric information. 
Specifically, the proposed framework, abbreviated as PointCG, consists of a \textbf{H}idden \textbf{P}oint \textbf{C}ompletion~(HPC) module and an \textbf{A}rbitrary-view \textbf{I}mage \textbf{G}eneration~(AIG) module. 
We first capture visible points from arbitrary views as inputs by removing hidden points. Then, HPC extracts representations of the inputs with an encoder and completes the entire shape with a decoder, while AIG is used to generate rendered images based on the visible points' representations. 
Extensive experiments demonstrate the superiority of the proposed method over the baselines in various downstream tasks. Our code will be made available upon acceptance.
\end{abstract}

}
\maketitle

\section{Introduction} 
Self-supervised representation learning (SSRL) aims to fully exploit the statistical and structural knowledge inherent in unlabeled datasets, enabling the encoder of the pre-training model to extract informative and discriminative representations.
The pre-trained encoder can be subsequently applied to various downstream tasks such as classification, segmentation, and object detection~\cite{PointGame_2023, gulipeng}. 
The core of SSRL lies in the design of appropriate pretext tasks aimed at aiding the encoder in achieving a full perception and understanding of the inputs. 

\begin{figure}[htbp]
    \centering
    \includegraphics[width=1.0\linewidth]{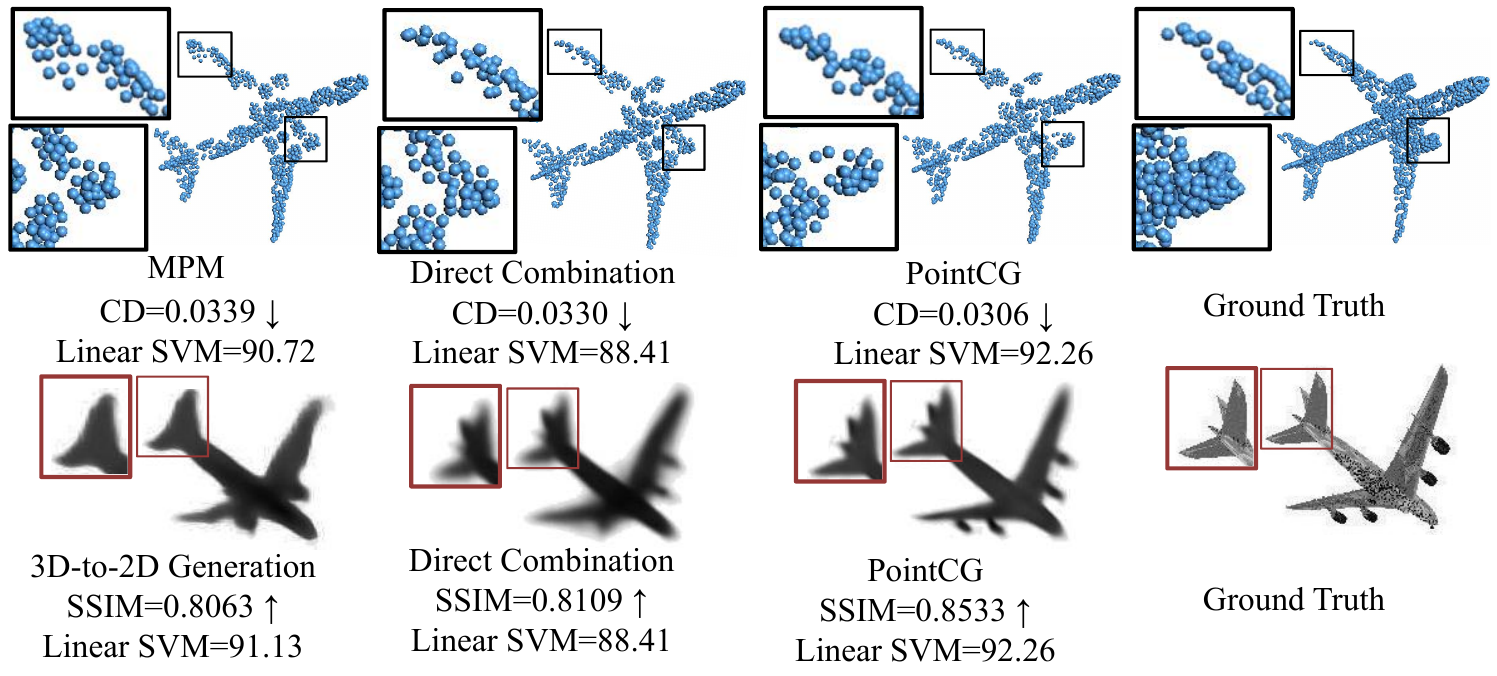}
    \caption{
    Qualitative and quantitative comparison of models using different pretext tasks.
    Chamfer Distance~(CD) and Structural Similarity Index~(SSIM) are employed as the quantitative metrics.
    For the masked point modeling~(MPM) task, we utilize the method proposed in Point-MAE~\cite{pang2022masked} with the inputs of visible points from arbitrary views (see Sec. \ref{sec:data_Org}). 
    For the 3D-to-2D generation task, we define the pretext task as generating images from arbitrary views. 
    The result of the model using only MPM exhibits group clustering at the edges, while our method yields sharpened and clear edges that closely align with the ground truth. 
    The model relying solely on 3D-to-2D generation fails to capture three-dimensional structural information, while our method can effectively preserve the geometric structure. 
    Directly combining both tasks generates point clouds and images superior to using only MPM or 3D-to-2D generation (Direct Combination) but with lower Linear-SVM accuracy.
    }
    \label{fig:motivation_intro}
\end{figure}

Based on the tasks employed, existing self-supervised pre-training methods can be broadly classified into two paradigms: contrastive learning and generative learning, both of which have attained great success in processing 2D images~\cite{he2022masked, he2020momentum, Chen_2021_SimSiam} and 3D point clouds~\cite{yu2022pointBERT, pang2022masked, zhang2022pointM2AE, huang2021STRL, wang2021OCCO, zhang2023pointvst}. 
Compared to contrastive learning, generative learning is considered as a more data-efficient pre-training method, capable of capturing the patterns of the inputs with relatively limited data volume~\cite{Recon_ICML_2023_Yili}. Therefore, it is highly favored in the context of data scarcity within the field of 3D vision, where masked point modeling~\cite{yu2022pointBERT, pang2022masked, zhang2022pointM2AE, jiang2022masked, wang2021OCCO} and 3D-to-2D generation~\cite{zhang2023pointvst, Wang_2023_TAP} stand out as two representative generative learning methods.
Among them, masked point modeling drives the model to predict arbitrary missing parts based on the remaining points. Accomplishing this task requires a thorough understanding of the spatial properties and global-local context of point clouds. 
3D-to-2D generation employs a cross-modal pretext task which translates a 3D object point cloud to its diverse forms of 2D rendered images~(e.g., silhouette, depth, contour). Pre-training with pixel-wise precise supervision drives the backbone to perceive the fine-grained edge details of 3D objects. 

\par However, both of the above methods have their own limitations. 
\textcolor{black}{As revealed in~\cite{Feng_CVPR2021_Recurrent, Sauder_NeurIPS2019_SelfSup, Zeng_CVPR2021_CorrNet3D}, due to the irregularity of point clouds, commonly used point set similarity metrics~(e.g., Chamfer Distance and Earth Mover’s Distance) in masked point modeling cannot provide explicit point-to-point supervision between ground truth and generated point clouds. The lack of precise correspondence results in limited feature representation capability of the pre-trained backbone network. }
Conversely, 3D-to-2D generation~\cite{zhang2023pointvst, Wang_2023_TAP} alleviates the issue of insufficient supervision signals by utilizing regular 2D images as the generation objective, offering pixel-wise precise supervision. 
However, relying solely on images from limited views as ground truth may overlook the structural information from occluded point sets, diminishing the backbone's perception of the spatial properties of point clouds. 
As shown in Fig.~\ref{fig:motivation_intro}, masked point modeling exhibits subpar performance in reconstructing some challenging areas (e.g., edges) due to the lack of point-to-point supervision. Besides, 3D-to-2D generation yields images lacking three-dimensional structural information, attributed to the lack of explicit geometric guidance. These observations collectively indicate the models' inadequate perception of the inputs, consequently reducing their performance on downstream tasks. 



\par Based on the aforementioned analysis, an intuitive method is to combine these two pretext tasks to retain their individual merits while compensating for their respective limitations. 
However, as shown in Fig. \ref{fig:motivation_intro}, while the model directly combining both tasks outperforms those relying solely on MPM or 3D-to-2D generation in generating high-quality point clouds or images, its Linear-SVM accuracy is lower ($88.41\%$ vs $90.72\%$ and $91.13\%$). 
We argue that the encoder's involvement in both tasks can lead to confusion when generating content for two modalities concurrently. Furthermore, to accomplish both tasks, the model shifts its training focus toward the decoder, which is typically discarded after pre-training. This phenomenon diminishes the feature extraction capability of the encoder, ultimately reducing the Linear SVM accuracy.

\begin{figure}[htbp]
    \centering
    \includegraphics[width=1.0\linewidth]{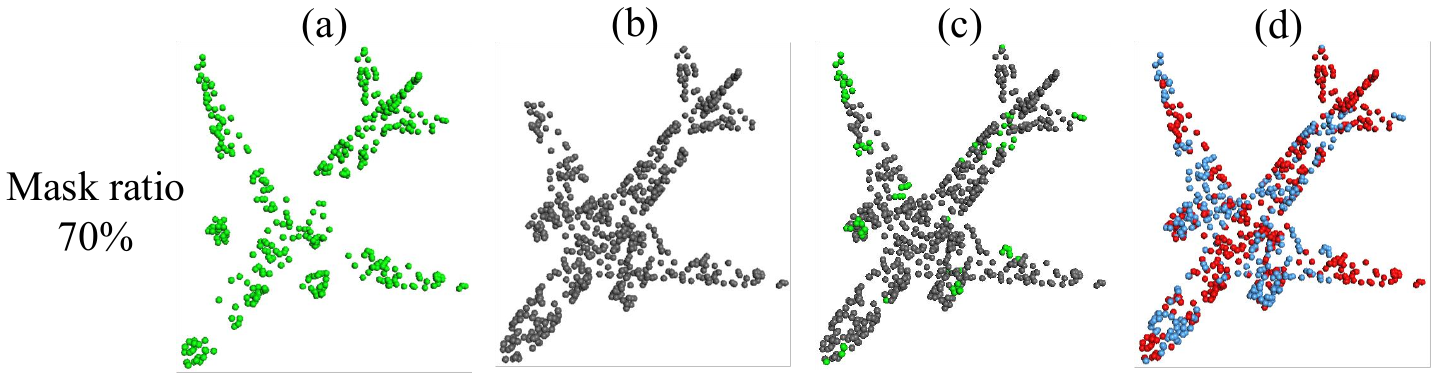}
    \caption{
    \textcolor{black}{Visualization of the unmasked points~(a), the masked points~(b), the completed point cloud composed of green unmasked points and gray masked points (c), and the completed point cloud in blue with overlapping points highlighted in red (d).}
    }
    \label{fig:MAE_OverAll_Overlaps}
\end{figure}
To address these issues, we propose PointCG, a framework that effectively integrates masked point modeling and 3D-to-2D generation tasks. This framework incorporates a \textbf{H}idden \textbf{P}oint \textbf{C}ompletion~(HPC) module and an \textbf{A}rbitrary-view \textbf{I}mage \textbf{G}eneration~(AIG) module. 
\textcolor{black}{Existing MAE-based MPM methods often employ a random masking strategy based on Farthest Point Sampling (FPS) and K-Nearest Neighbor (KNN) techniques. }
\textcolor{black}{However, the inputs of unmasked patches~(Fig.~\ref{fig:MAE_OverAll_Overlaps} (a)) preserve the overall shape of an object and exhibit substantial overlap with the target points~(highlighted in red in Fig.~\ref{fig:MAE_OverAll_Overlaps} (d)). 
The leakage of overall structure and point location information enables the model to reconstruct the object without a holistic comprehension of the entire structure, which limits the learning capacity of the encoder during pre-training. 
}
To overcome this limitation, we select the visible points from arbitrary views by removing hidden points as input and introduce the HPC module to complete the point clouds. 
For the 3D-to-2D generation task, we employ the arbitrary-view image generation as the pretext task, which generates the image from an arbitrary view based on the representations of visible points extracted by the encoder. 
\textcolor{black}{Furthermore, the cross-modal feature alignment is introduced to align the feature spaces of point clouds and images, which enables simultaneous content generation across both modalities and refocuses the training on the encoder.}
\textcolor{black}{
Specifically, we extract features from both the input point clouds and their corresponding rendered 2D images, encouraging feature proximity for the same instance while maintaining feature separation for different instances. 
}

\par Through the effective integration of HPC and AIG, the pre-trained encoder achieves a comprehensive understanding of 3D objects and can extract high-quality 3D representations. 
We evaluate our model and the proposed modules with a variety of downstream tasks and ablation studies. We further demonstrate that informative representations can be effectively learned from the restricted points, and such representations facilitate effortless masked point modeling and arbitrary-view image generation.

\IEEEpubidadjcol

\section{Related Work} \label{sec:relatedWork}
\subsection{Self-supervised Representation Learning}
Self-supervised representation learning aims to derive robust and general representations from unlabeled datasets, which can be broadly classified into two categories based on the types of pretext tasks: contrastive learning and generative learning.

Contrastive learning-based methods (e.g., BYOL~\cite{grill2020bootstrap}, SimSiam~\cite{Chen_2021_SimSiam}, DINO~\cite{Caron_2021_DINO}, STRL~\cite{huang2021STRL}, CrossPoint~\cite{afham2022crosspoint}) define the augmented views of a sample as positive samples, while considering other instances as negative samples, thereby constructing discriminative tasks. 
Generative learning-based methods (e.g., GPT~\cite{radford2019language}, Point-BERT~\cite{yu2022pointBERT}, Point-MAE~\cite{pang2022masked}, OcCo~\cite{wang2021OCCO}, MaskFeat3D~\cite{MaskFeat3D_Yan_2024_ICLR}) are based on the intuition that effective feature abstractions contain sufficient information to reconstruct the original geometric structures~\cite{zhang2023pointvst}. 
In the point cloud processing community, where 3D assets are relatively scarce, generative learning has garnered widespread attention due to its data efficiency~\cite{he2022masked, pang2022masked, jiang2022masked}. Among them, MAE stands out as one of the representative paradigms. 
It involves masking a substantial portion of input data, followed by the use of an encoder to extract informative representations and a decoder to reconstruct explicit features (e.g., pixels or points) or implicit features (e.g., discrete tokens). 
\textcolor{black}{
Taking Point-BERT~\cite{yu2022pointBERT}, MaskFeat3D~\cite{MaskFeat3D_Yan_2024_ICLR}, and IAE~\cite{Yan_2022_IAE} as examples, each of these methods utilizes the visible groups as input after masking and reconstructs the positions of masked points, surface normals, and surface variations, as well as the implicit features of the masked points.
However, after random masking~\cite{pang2022masked} or partial occlusion~\cite{jiang2022masked}, the visible groups often retain the overall structure of the object~(Fig.~\ref{fig:PartRecon_MN40}), and there are substantial overlap regions between input and target patches~(Fig.~\ref{fig:MAE_OverAll_Overlaps}).
The leakage of overall structure and point location information will reduce the difficulty in reconstructing masked patches, thus limiting the learning and inference capabilities of the encoder. }

\par To avoid the leakage of the object's overall shape and minimize overlap, we simulate scanners to capture visible points from arbitrary views as input.
\textcolor{black}{Our approach is conceptually aligned with OcCo~\cite{wang2021OCCO}, which employs the Z-Buffer algorithm~\cite{straber1974schnelle} to select visible points from multiple views, subsequently completing the original point clouds with an encoder-decoder architecture. }
\textcolor{black}{
The Z-Buffer algorithm addressed within rendering relies on two assumptions: the points satisfy sampling criteria (e.g., Nyquist condition) and the points are associated with normals (or the normals can be estimated)~\cite{katz2007direct}.
However, our method seeks rigorous theoretical support for visibility computation without requiring normal estimation, point rendering, or surface reconstruction. Therefore, we employ the Hidden Point Removal~(HPR) operator to compute visibility in a more robust manner.
}

\subsection{Cross-modal Learning} 
Recently, cross-modal learning has been a popular research topic, aiming at extracting informative representations from multiple modalities such as images, audio, and point clouds. It has the potential to enhance the performance of various tasks, including visual recognition, speech recognition, and point cloud analysis.

In point cloud analysis, a variety of methods have been proposed for cross-modal learning, such as CrossPoint~\cite{afham2022crosspoint}, PointMCD~\cite{zhang2023pointmcd}, TAP~\cite{Wang_2023_TAP}, and PointVST~\cite{zhang2023pointvst}. 
\textcolor{black}{CrossPoint~\cite{afham2022crosspoint} establishes cross-modal contrastive learning between images and point clouds, demonstrating that the correspondence between images and points can enhance 3D object understanding. }
\textcolor{black}{PointMCD~\cite{zhang2023pointmcd} obtains a powerful point encoder by aligning the multi-view visual and geometric descriptors generated by a pretrained image encoder and a learnable point encoder.
Both CrossPoint~\cite{afham2022crosspoint} and PointMCD~\cite{zhang2023pointmcd} are based on the contrastive paradigm and rely heavily on extensive 3D-2D paired data.} 
\textcolor{black}{
Generative methods, such as TAP~\cite{Wang_2023_TAP} and PointVST~\cite{zhang2023pointvst}, generate images from specific views based on the input point clouds. These methods use regular 2D images as generation objectives to provide precise supervision. }

\par \textcolor{black}{In this paper, we follow the generative learning paradigm and propose a unified pre-training framework with two complementary pretext tasks: hidden point completion and arbitrary-view image generation. 
The spatial awareness provided by 3D completion addresses the geometric insensitivity inherent in image supervision, as shown in the second line of column one in Fig.~\ref{fig:motivation_intro}. }
Additionally, we demonstrate the mutual enhancement between the two pretext tasks through various experiments. 

\section{Methodology} \label{sec:Method}
\begin{figure*}[htbp]
    \centering
    \includegraphics[width=0.9\linewidth]{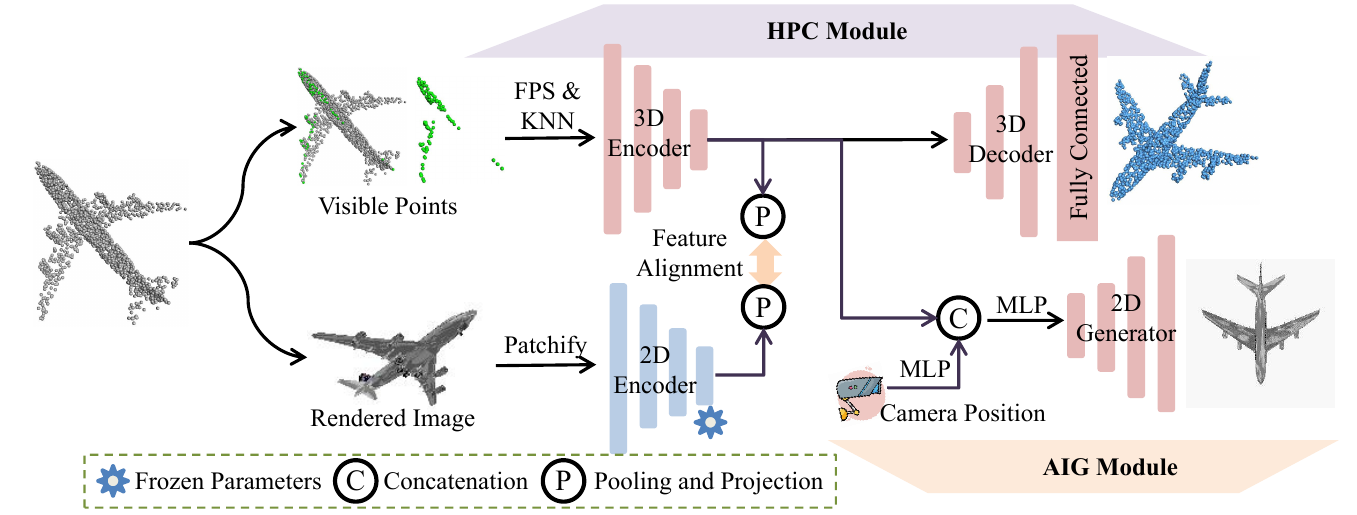}
    \caption{
    Overview of PointCG. PointCG integrates two prevalent methods, masked point modeling~(MPM) and 3D-to-2D generation, as pretext tasks within a pre-training framework. In detail, we first capture visible points with the HPR~\cite{katz2007direct} operator. Then we utilize an encoder-decoder architecture to extract features from these visible points and complete the original point clouds through the hidden point completion~(HPC) module. 
    The arbitrary-view image generation~(AIG) module generates images based on the aligned representations of visible points. 
    Note that the input images for feature alignment are randomly selected and do not need to match the target images of image generation.
    }
    \label{fig:OverView}
\end{figure*}

\par As illustrated in Fig.~\ref{fig:OverView}, PointCG mainly consists of a hidden point completion~(HPC) module and an arbitrary-view image generation~(AIG) module. 
Specifically, we begin by selecting the visible points from arbitrary views as the inputs (Sec.~\ref{sec:data_Org}), 
and then introduce an asymmetric Transformer-based encoder-decoder architecture for extracting representations and completing hidden points (Sec. \ref{sec:OcCo}). 
Finally, we generate arbitrary-view images (Sec. \ref{sec:CM_Guide}) based on the aligned representations extracted by the encoder. 
In the following, we will delve into the details of these modules.


\subsection{Data Organization} \label{sec:data_Org}
\par Given a complete point cloud $P=\{p_{i}|{1}\leq{i}\leq{N}\} \in \mathbb{R}^{3}$, we randomly select the camera position $C=[azimuth, elevation, distance]$, where $distance$ is fixed at 1.0. $azimuth$ and $elevation$ are randomly chosen within the range of $[0, 2\pi]$.
The HPR operator~\cite{katz2007direct} is employed to determine whether ${p_i}$ is visible from $C$. It mainly consists of two steps: inversion transformation and convex hull construction. 
\par
\begin{figure}[htbp]
    \centering
    \includegraphics[width=0.90\linewidth]{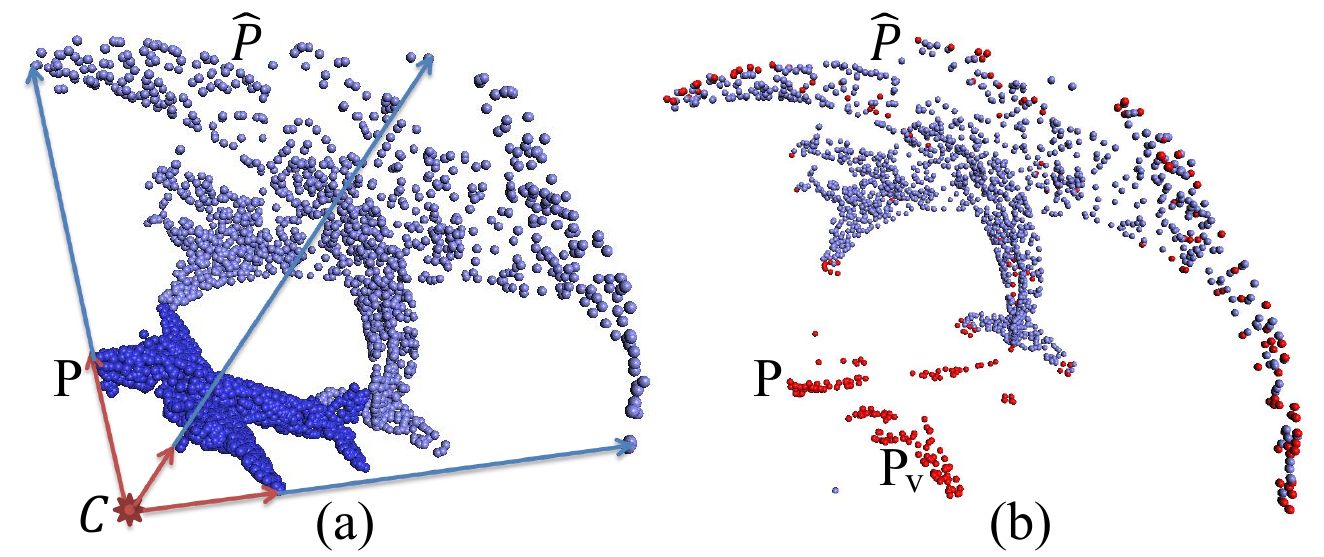}
    \caption{Visualization of the original points ${P}$ in blue, points after spherical flipping $\hat{P}$ in light blue and visible points from $C$ in magenta.}
    \label{fig:SphericalFlip}
\end{figure}

 \textbf{Inversion transformation}. 
We employ spherical flip~\cite{katz2005mesh} to reflect each point $p_i\in{P}$ to the spherical surface~(denoted as $\hat{p_i}\in\hat{P}$) along the ray from $C$ through $p_i$ to the spherical surface, as illustrated in Fig.~\ref{fig:SphericalFlip}~(a). 
\par \textbf{Convex hull construction}. 
The visible points from $C$ inverted on the spherical surface are situated on the convex hull of $\hat{P}\cup{C}$. 
Therefore, we need to compute the collection of triangular planes, which make up the convex hull. Then we extract all vertices~(magenta points in $\hat{P}$ in Fig.~\ref{fig:SphericalFlip}~(b)) of the convex hull and project them back onto the original point cloud to obtain the visible points $P_v$~(magenta points in $P$ in Fig.~\ref{fig:SphericalFlip}~(b)). The remaining points of the original point cloud are hidden from $C$, denoted as $P_h$. 
\subsection{Occluded Point Completion} \label{sec:OcCo}

For each input, we employ the FPS and KNN to divide the visible points $P_v$ into patches with $v$ centers. Simultaneously, we extract $h$ central points from the hidden points $P_h$ and retrieve $k$ nearest neighbor points from the complete point cloud as the target patches $P_{GT}$. 
Then, the visible patches are projected into tokens $T^v\in\mathbb{R}^{v\times{d}}$ with a lightweight PointNet~\cite{qi2017pointnet}, where $d$ is the dimension of features. 
Subsequently, a learnable Multi-Layer Perceptron (MLP) is adopted to embed the visible and hidden centers into positional tokens denoted as $T_L^{v}$ and $T_L^{h}$, respectively. 
Finally, we extract representations $T_E$ by an encoder and capture the tokens $T_D$ with a decoder for completing the original point clouds:
\begin{equation}
\begin{split}
T_E=Encoder(T^v,T_L^{v}),  T_E\in\mathbb{R}^{v\times{g}\times{d}}
\label{eq:eq_Encoder}
\end{split}
\end{equation}
\begin{equation}
\begin{split}
T_D=Decoder(Cat(T_E,T_H),Cat(T_L^{v},T_L^{h})) \\T_H\in\mathbb{R}^{h\times{d}},  T_D\in\mathbb{R}^{(v+h)\times{d}}
\label{eq:eq_Decoder}
\end{split}
\end{equation}
\textcolor{black}{where $T_H$ represents the hidden tokens, which is initialized by duplicating a learnable token of dimension $d$. }
We concatenate the visible points' features $T_E$ and $T_H$, as well as the positional tokens $T_L^{v}$ and $T_L^{h}$ as the inputs of the decoder. 
Based on the outputs $T_D$ of the decoder, we will reconstruct the $k$ nearest neighbors of $h$ center points by a reconstruction head of a fully connected~(FC) layer:
\begin{equation}
\label{eq:Rebuild_HidPoints}
\begin{split}
P_{pre}=Reshape(FC(T_D)), \\ P_{pre}\in\mathbb{R}^{h\times{k}\times{3}}
\end{split}
\end{equation}
where $P_{pre}$ denotes as the predicted hidden point patches.

\par \textbf{Loss function}. The Chamfer distance~\cite{fan2017point} is employed as the reconstruction loss: 
\begin{equation}
\begin{split}
\mathcal{L}_{CD}=\frac{1}{|P_{pre}|}\sum_{\hat{x}\in{P_{pre}}}\min_{x\in{P_{GT}}}||\hat{x}-x||_2^2 \\ +\frac{1}{|P_{GT}|}\sum_{x\in{P_{GT}}}\min_{\hat{x}\in{P_{pre}}}||\hat{x}-x||_2^2
\label{eq:loss_Rebuild}
\end{split}
\end{equation}
where $P_{GT}\in\mathbb{R}^{h\times{k}\times{3}}$ denotes the reconstruction targets. 

\subsection{Arbitrary-view Image Generation} \label{sec:CM_Guide}
\subsubsection{Feature Alignment}
To shift the pre-training focus towards enhancing the encoder for better 3D understanding, we employ the feature alignment module to build correspondence between images and point clouds in the feature space.


\par During pre-training, a pre-trained CLIP-visual~\cite{radford2021learning} module~$f$ is used to extract features from the rendered image $I_i$. 
Then, the image features $f(I_i)$ and \textcolor{black}{the 3D features $T_i\in{Max(T_E)}$} are projected into the invariant space with functions $g$ and $h$, respectively, resulting in $\mathcal{H}_i=g(f(I_i))$ and $\mathcal{Z}_i=h(T_i)$. 

\par \textbf{Loss function}. In the invariant space, we aim to maximize the similarity between $\mathcal{Z}_i$ and $\mathcal{H}_i$ when they correspond to the same objects. The cross-modal instance discrimination loss $\mathcal{L}_{CM}$ can be formulated as:  
\begin{equation}
\label{eq:loss_CrossModal}
\begin{split}
\begin{array}{l}
\mathcal{L}_{CM}=\frac{1}{2M}\sum^{M}_{i=1}[l(i,\mathcal{Z},\mathcal{H})+l(i,\mathcal{H},\mathcal{Z})] \\
\\
\textcolor{black}{l(i,\mathcal{Z},\mathcal{H})=}\\ 
\textcolor{black}{-log\frac{exp(s(\mathcal{Z}_i,\mathcal{H}_i)/{\tau})}{\sum^{M}_{k=1,k\neq{i}}{exp(s(\mathcal{Z}_i,\mathcal{Z}_k)/{\tau})}+\sum^{M}_{k=1}{exp(s(\mathcal{Z}_i,\mathcal{H}_k)/{\tau})}}
}
\end{array}
\end{split}
\end{equation}
where $M$ is the mini-batch size. $\tau$ is the temperature co-efficient, and $s(\centerdot)$ denotes the cosine similarity function. 

\subsubsection{Image Generation} \label{sec:Trans3Dto2D}
\par AIG generates rendered images from arbitrary views based on the visible points' representations extracted by the encoder.
\par In pre-training, we randomly select a rendered image as the target and capture the corresponding view parameters $L_C$ as the input. 
$L_C$ comprises azimuth $\phi$, elevation $\lambda$, and distance $\rho$, described as $L_C=Cat(\phi, \lambda, \rho)$. 
To enhance flexibility, we apply several learnable transformation layers for positional embedding tokens $T^C_L$. 
\textcolor{black}{Then, we concatenate $T_E$ with $T^C_L$ and encode the combined representation using MLP~($\mathcal{G}_{\theta}$): $T^I_G=\mathcal{G}_{\theta}Cat(T_E, T^C_L)$. }
\begin{figure}[htbp]
    \centering
    \includegraphics[width=0.95\linewidth]{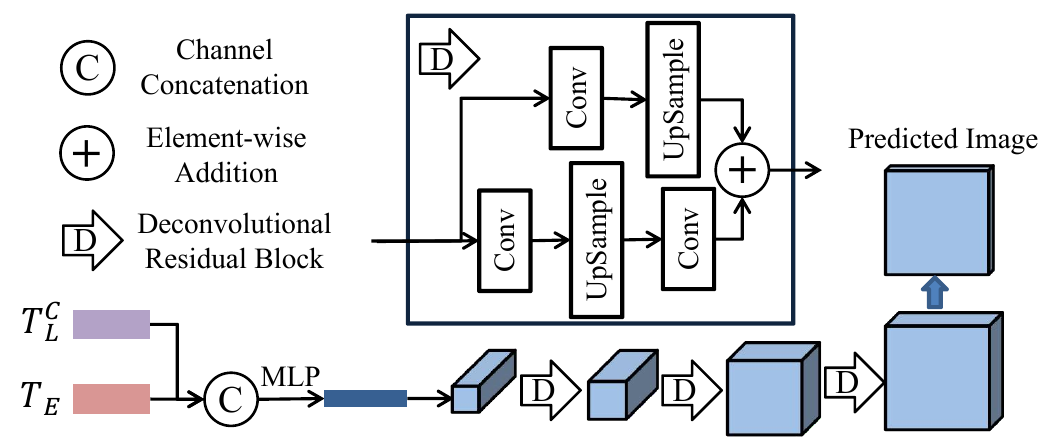}
    \caption{
    The image generator consists of several deconvolutional residual blocks, generating the image from the view of $L_C$. 
    }
    \label{fig:generateImgDetail}
\end{figure}
\textcolor{black}{Finally, we design an image generator~(Fig.~\ref{fig:generateImgDetail}) to generate rendered images based on $T^I_G$.}
\par Specifically, we start by reshaping $T^I_G$ from its original vectorized representation to a 2D feature map. 
This feature map is then passed through a series of deconvolutional residual blocks and three parallel convolutional blocks to generate the rendered image $G_I^p$.

\par \textbf{Loss function}. 
We utilize the $\mathcal{L}_1$ loss as the content loss for image generation:
\begin{equation}
\label{eq:loss_l1}
\begin{split}
\begin{array}{l}
\mathcal{L}_1={\frac{1}{n}}*\sum_{i=1}^n{|GT_I-G_I^p|}
\end{array}
\end{split}
\end{equation}
where $G_I^p$ and $GT_I$ represent the predicted and GT images, respectively, while $n$ denotes the number of sample points.
Besides, we incorporate multi-scale frequency reconstruction~(MSFR) loss~\cite{Sung2021_MIMO_UNet_MSFRloss} as the auxiliary loss alongside the content loss $\mathcal{L}_1$ to reduce the differences in the frequency space. MSFR loss measures the $\mathcal{L}_1$ distance between multi-scale GT and predicted images in the frequency domain:
\begin{equation}
\label{eq:loss_MSFR}
\begin{split}
\begin{array}{l}
\mathcal{L}_{MSFR}=\frac{1}{n}*\sum_{i=1}^{n}{|\mathcal{F}(GT_I)-\mathcal{F}(G_I^p)|}
\end{array}
\end{split}
\end{equation}
where $\mathcal{F}$ denotes the fast Fourier transform~(FFT) that transfers the image signal to the frequency domain. 
MSFR can effectively maintain contrast in high-frequency regions, complementing the ability of $\mathcal{L}_1$ to preserve colors and luminance~\cite{Zhao_2017_Losses}. 
The image generation loss is given by:
\begin{equation}
\label{eq:loss_GenImg}
\begin{split}
\begin{array}{l}
\mathcal{L}_G=\alpha*\mathcal{L}_1+\beta*\mathcal{L}_{MSFR} 
\end{array}
\end{split}
\end{equation}
where the contribution of $\mathcal{L}_1$ and $\mathcal{L}_{MSFR}$ can be adjusted by modifying the values of $\alpha$ and $\beta$.

Our loss function during pre-training is formulated  as:
\begin{equation}
\label{eq:loss_Overall}
\begin{split}
\begin{array}{l}
\mathcal{L}=\omega * \mathcal{L}_{CD}+\phi * \mathcal{L}_{CM}+\psi * \mathcal{L}_G
\end{array}
\end{split}
\end{equation}
where $\mathcal{L}_{CD}$ enforces 3D completion, $\mathcal{L}_{CM}$ introduces 3D-2D correspondence, and $\mathcal{L}_G$ ensures image generation.

\section{Experiment} \label{sec:Experiments}
\par In this section, we present extensive experiments to demonstrate the effectiveness of our method. We begin by introducing the pre-training process on ShapeNet55~\cite{chang2015shapenet}. Then, we showcase its performance on 3D completion tasks in Sec.~\ref{sec:Completion}.
Then, in Sec.~\ref{sec:ShapeClassification}, Sec.~\ref{sec:PartSeg}, and Sec.~\ref{sec:SemSeg}, we follow the previous works to conduct experiments of object classification, part segmentation, and semantic segmentation.
Finally, we validate the effectiveness of our modules through various ablation studies in Sec.~\ref{sec:Ablation}. 
\par In the following tables, $``\text{Pre-T}"$ indicates whether the model is initialized with a pre-trained model, while $``\text{Rep.}"$ signifies that the result is reproduced with the official code. 
Please note that we reproduce experiments of Point-MAE~\cite{pang2022masked} and Point-M2AE~\cite{zhang2022pointM2AE} with their official codes, and all settings are consistent with our experimental configuration. 

\par \textbf{Pre-training}. We pre-train the encoder on ShapeNet55~\cite{chang2015shapenet}, which contains $52,470$ clean 3D models, covering $55$ common object categories. The input point number $N$ is $2,048$, and the rendered images have a size of $224\times{224}\times{3}$. 
The encoder and decoder include $12$ and $4$ standard Transformer blocks, respectively. 
Each Transformer block has $384$ hidden dimensions with $6$ heads. We employ the AdamW optimizer~\cite{loshchilov2017decoupled} and cosine learning rate decay~\cite{loshchilov2016sgdr}. The initial learning rate is set to $0.001$, and the weight decay is $0.05$. 

\subsection{3D Completion}  \label{sec:Completion}
\par
\begin{figure}[htbp]
    \centering
    \includegraphics[width=1.0\linewidth]{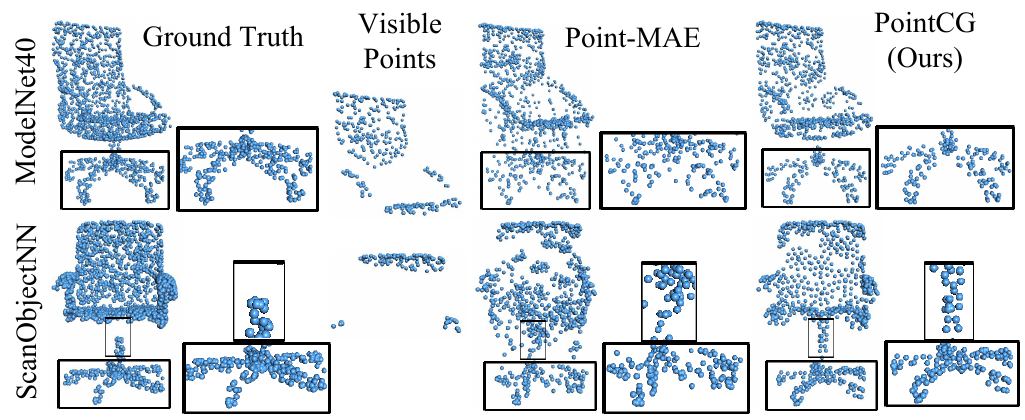}
    \caption{
    3D completion with pre-trained models of Point-MAE~\cite{pang2022masked} and our method, based on the unmasked points from ModelNet40~\cite{wu2015_modelNet} (top row) and ScanObjectNN~\cite{uy2019revisiting} (bottom row). 
}
    \label{fig:OclusionRecon_MN40_ScanObj_nobg}
\end{figure}

 \textbf{Completion based on visible points from random views}. 
To assess the effectiveness of our self-supervised model initialized with pre-trained weights, we randomly select an instance from synthetic dataset ModelNet40~\cite{wu2015_modelNet} and real-world dataset ScanObjectNN~\cite{uy2019revisiting} separately and reconstruct the original point clouds.
The visualization results of Point-MAE~\cite{pang2022masked} and our model are shown in Fig.~\ref{fig:OclusionRecon_MN40_ScanObj_nobg}. 

Compared to Point-MAE, our method not only completes the chair's pivot axis and the five-pronged base with greater fidelity but also obtains smoother surface structures. 
Our method achieves remarkable performance in reconstructing both synthetic and real-world data with visible points from arbitrary views. 


\par
\begin{figure}[htbp]
    \centering
     \includegraphics[width=1.0\linewidth]{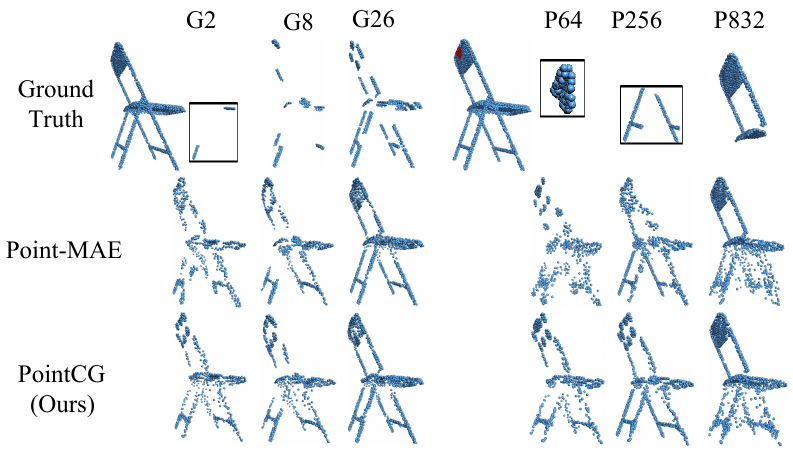}
    \caption{
    3D completion with pre-trained models of Point-MAE~\cite{pang2022masked} and our model based on partial points from ModelNet40~\cite{wu2015_modelNet}. 
    }
    \label{fig:PartRecon_MN40}
\end{figure}

 \textbf{Completion based on grouped patches and partial points}. To demonstrate the robustness and generalizability of our method in the 3D completion task, we devise two methods to obtain grouped patches and partial points as inputs. The reconstruction results are visualized in Fig.~\ref{fig:PartRecon_MN40}. 
In the first method, we obtain $2$, $8$, and $26$ central points via FPS, and then acquire $32$ neighboring points with KNN to form groups~\cite{pang2022masked, zhang2022pointM2AE} denoted by `G2', `G8', and `G26'. 
In the second method, we randomly select one group, consisting of $64$, $256$, and $832$ points, denoted as `P64', `P256', and `P832', respectively.

For `G2' and `G8', both Point-MAE and our method can complete the overall structure, but our method excels in recovering finer geometric details and sharper edges.
The `G26' column retains sufficient information, and the results are satisfactory for both methods. 
For `P64' and `P256', our method successfully completes entire structures and captures many local details. In contrast, the results of Point-MAE appear quite blurry. 
Although the `G26' and `P832' have the same number of input points, the reconstruction results of Point-MAE based on `P832' are significantly lower than `G26'. 
\par \textcolor{black}{
As shown, the inputs obtained by random masking retain objects' structural information while exposing the coordinates of target points. These MAE methods employing random masking strategy, such as Point-MAE, exhibit poorer reconstruction performance when the inputs are partial and lack of complete structural integrity. 
}
Our method, however, excels in extracting informative representations and demonstrates strong inference capability, leading to superior reconstruction performance, even from highly partial point data.

\subsection{Shape Classification} \label{sec:ShapeClassification}
\par 
To assess the discrimination of the representations extracted by the pre-trained encoder, we validate the encoder on the shape classification task using the ModelNet40~\cite{wu2015_modelNet} and ScanObjectNN~\cite{uy2019revisiting} datasets. 

\par \textbf{Shape classification on synthetic data}.
ModelNet40~\cite{wu2015_modelNet} contains $12,311$ clean 3D CAD models, covering $40$ object categories. 
We fine-tune the pre-trained encoder, and the results are presented in Tab.~\ref{tab:shapeClsMN40_GL_MLP3}.
Our method achieves 94.03\% with global fine-tuning, surpassing the reproduced version of Point-MAE~(Rep.) (93.21\%) by 0.82\% and the publicly released accuracy by 0.23\%. 
To validate the effectiveness of our architecture, we incorporate Point-M2AE as the backbone and evaluate its classification performance on ModelNet40. 
The classification results outperform the outcomes of the reproduced Point-M2AE model.
\begin{table}[htbp]
\caption{ 
Accuracy of shape classification on ModelNet40. 
}
\centering
\begin{tabular}{lcccc}
\hline
\centering  
Methods                         & Pre-T  & booktitle/year. & Acc~(Vote). \\ \hline
DGCNN~\cite{wang2019DGCNN}               & -     & ACM/2019    & 92.9  \\
RSCNN~\cite{liu2019relation}             & -     & CVPR/2019   & 93.6 \\ 
PointTransformer~\cite{zhao2021point}    & -     & ICCV/2021     & 93.7 \\
DGCNN+OcCo~\cite{wang2021OCCO}          & Y     & ICCV/2021    & 93.0 \\
DGCNN+STRL~\cite{huang2021STRL}         & Y     & ICCV/2021    & 93.1 \\
DGCNN+MAE3D~\cite{jiang2022masked}      & Y     & TMM/2023     & 93.4  \\
Point-BERT~\cite{yu2022pointBERT}       & Y     & CVPR/2022    & 93.2  \\ 
MaskPoint~\cite{Liu_ECCV_MaskPoint}     & Y     & ECCV/2022    & 93.8  \\
Point-MAE~\cite{pang2022masked}         & Y     & ECCV/2022    & 93.8 \\ 
Joint-MAE~\cite{guo2023joint}           & Y     & CoRR/2023    & 94.0 \\
Point-MAE~(Rep.)                        & Y     & ECCV/2022    & 93.21 \\
\textcolor{black}{PointCG~(Point-MAE)}    & Y     & -     & \textbf{94.03} \\ 
Point-M2AE~\cite{zhang2022pointM2AE}    & Y     & NeurIPS/2022 & 94.0  \\ 
Point-M2AE~(Rep.)                      & Y     & NeurIPS/2022    & 93.59 \\
PointCG~(Point-M2AE)                 & Y     & -     & 94.11 \\ 
\bottomrule
\end{tabular}
\label{tab:shapeClsMN40_GL_MLP3}
\end{table}

\par Besides, we also attempt to freeze the parameters of our pre-trained model, and validate it with a Linear-SVM classifier in Tab.~\ref{tab:MN40_Cls_SVM}. 

\textcolor{black}{
Our method outperforms Point-MAE~\cite{pang2022masked} and Point-M2AE~\cite{zhang2022pointM2AE} by margins of +1.46\% and +0.29\%, respectively. }
The results highlight the superior quality of the 3D representation learned by our method.
\begin{table}[htbp]
\caption{Accuracy of Linear-SVM on ModelNet40. 
}
\centering
\begin{tabular}{lcc}
\hline
\centering
Methods                                   & Pre-T       & Acc.  \\
\hline
FoldingNet~\cite{yang2018foldingnet}      & -      & 88.4  \\
DGCNN+Jiasaw~\cite{sauder2019self}        & Y          & 90.6  \\
DGCNN+OcCo~\cite{wang2021OCCO}            & Y         & 89.2  \\
DGCNN+CrossPoint~\cite{afham2022crosspoint} & Y        & 91.2  \\ 
\textcolor{black}{FoldingNet+PointMCD~\cite{zhang2023pointmcd}} & Y    & 89.8 \\
Point-BERT~\cite{yu2022pointBERT}    & Y     & 87.4  \\
Joint-MAE~\cite{guo2023joint}       & Y       & 92.4  \\
Point-MAE~\cite{pang2022masked}     & Y      & 91.0  \\
PointCG~(Point-MAE)             & Y         & \textbf{92.26} \\ 
Point-M2AE~\cite{zhang2022pointM2AE}     & Y    & 92.9  \\
Point-M2AE~(Rep.)                        & Y     & 92.63  \\
PointCG~(Point-M2AE)              & Y      & \textbf{92.92} \\ 
\bottomrule
\end{tabular}
\label{tab:MN40_Cls_SVM}
\end{table}

\textbf{Shape classification on the real-world data}. 
Evaluating a pre-trained model's performance on real-world datasets is crucial, as real-world scenes tend to be more complex than synthetic ones. 
We follow the common practice to evaluate our model on three variants: `OBJ-BG', `OBJ-ONLY', and `PB-T50-RS' of ScanObjectNN~\cite{uy2019revisiting}.

\par \textcolor{black}{ 
To further validate the effectiveness of our design, we follow PointVST~\cite{zhang2023pointvst} with DGCNN as the encoder to construct a pre-training network~(DGCNN+PointCG) and evaluate it on the `PB-T50-RS' variant.
Additionally, we reproduce TAP~\cite{Wang_2023_TAP} and PointVST~\cite{zhang2023pointvst} using their official pretrained models. 
}
\par \textcolor{black}{As presented in Tab.~\ref{tab:Cls_ScanObjectNN_GlobalFT}, our method outperforms CrossPoint~\cite{afham2022crosspoint}, as well as the reproduced TAP~\cite{Wang_2023_TAP} and PointVST~\cite{zhang2023pointvst}, all using DGCNN~\cite{wang2019DGCNN} as the encoder. 
When utilizing Point-MAE or Point-M2AE as the backbone, the classification results exhibit a significant improvement over the reproduced Point-MAE and Point-M2AE. }
These results underscore the discriminative power of the representations extracted by the encoder, even in complex real-world scenes.
\begin{table}[htbp]
\caption{Accuracy of shape classification on ScanObjectNN. 
}
\centering
\begin{tabular}{lccc} 
\hline
Methods                                 & OBJ-BG & OBJ-ONLY & PB-T50-RS \\ \hline
DGCNN~\cite{wang2019DGCNN}        & 82.8   & 86.2   & 78.1 \\ 
DGCNN+MAE3D~\cite{jiang2022masked}  & 87.7   & 88.4    & 86.2   \\ 
\textcolor{black}{DGCNN+CrossPoint~\cite{afham2022crosspoint}}  & -  & -  & 86.2  \\
\textcolor{black}{DGCNN+TAP~\cite{Wang_2023_TAP}}  & -  & -  & 86.6  \\ 
\textcolor{black}{DGCNN+TAP~\cite{Wang_2023_TAP}(Rep.)}  & -  & -  & 86.54  \\ 
\textcolor{black}{DGCNN+PointVST~\cite{zhang2023pointvst}}  & -  & -  & 89.3  \\
\textcolor{black}{DGCNN+PointVST~\cite{zhang2023pointvst}~(Rep.)}  & -  & -  & 87.6  \\  
\textcolor{black}{DGCNN+PointCG}  & -  & -  & 87.90  \\ \hline
Transformer+OcCo~\cite{wang2021OCCO}  & 84.85  & 85.54    & 78.79  \\ 
Transformer+TAP~\cite{Wang_2023_TAP}  & 90.36  & 89.50   & 85.67  \\
Point-BERT~\cite{yu2022pointBERT}       & 87.43   & 88.12    & 83.07  \\ 
Joint-MAE~\cite{guo2023joint}         & 90.94  & 88.86   & 86.07  \\ 
Point-MAE~\cite{pang2022masked}     & 90.02   & 88.29    & 85.18  \\ 
PointCG~(Point-MAE)                  & 91.16  & \textbf{88.99}   & \textbf{86.47}  \\ 
Point-M2AE~\cite{zhang2022pointM2AE}   & \textbf{91.22}  & 88.81  & 86.43 \\ 
Point-M2AE~(Rep.)            & 90.87  & 88.12  & 85.39 \\ 
PointCG~(Point-M2AE)         & 91.19  & 88.72   & 86.41  \\ 
\bottomrule
\end{tabular}
\label{tab:Cls_ScanObjectNN_GlobalFT}
\end{table}

\par \textcolor{black}{ \textbf{Few-shot Learning.} Following previous works~\cite{sharma2020self, yu2022pointBERT, wang2021OCCO, pang2022masked}, we conduct few-shot learning experiments using the pre-trained model on ModelNet40~\cite{wu2015_modelNet}. We adopt $n$-way, $m$-shot setting, where $n$ denotes the number of classes randomly selected from the dataset, and $m$ represents the number of objects randomly sampled for each class. This yields ${n}\times{m}$ objects for training. For evaluation, we randomly select $20$ unseen objects from each of $n$ classes.}

\par \textcolor{black}{The results with settings of $n\in\{5,10\}$ and $m\in\{10,20\}$ are presented in Tab.~\ref{tab:Cls_MN40_FewShot}. 
As shown, our method consistently outperforms the baselines in nearly all few-shot settings, with minimal deviation. 
This highlights the robustness and generalization of the representations extracted by the PointCG encoder, even in data-limited scenarios.}

\begin{table*}[htpb]
\caption{\textcolor{black}{Few-shot object classification on ModelNet40. We conduct 10 independent experiments for each setting and report mean accuracy (\%) with standard deviation.}}
\centering
\begin{tabular}{lcccc}
\toprule
Methods                                 & 5-way, 10-shot   & 5-way, 20-shot  & 10-way, 10-shot  & 10-way, 20-shot \\ \hline
Transformer~\cite{vaswani2017attention} & $87.8 \pm 5.2$  & $93.3 \pm 4.3$  & $84.6 \pm 5.5$    & $89.4 \pm 6.3 $  \\
Transformer+OcCo~\cite{wang2021OCCO} & $94.0 \pm 3.6$  & $95.9 \pm 2.3$  & $89.4 \pm 5.1$    & $92.4 \pm 4.6 $  \\
Point-BERT~\cite{yu2022pointBERT}       & $94.6 \pm 3.1$  & $96.3 \pm 2.7$  & $91.0 \pm 5.4 $   & $92.7 \pm 5.1$ \\ 
Point-MAE~\cite{pang2022masked}   & $96.3 \pm 2.5$  & $97.8 \pm 1.8$  & $92.6 \pm 4.1$   & 95.0 $\pm$ 3.0 \\ 
PointCG~(Point-MAE)       & \textbf{96.7} $\pm$ 2.1  & \textbf{98.0} $\pm$ 1.3  & \textbf{93.1} $\pm$ 3.6    & \textbf{95.8} $\pm$ 2.6 \\ 
Point-M2AE~\cite{zhang2022pointM2AE}   & 96.8 $\pm$ 1.8  & 98.3 $\pm$ 1.4  & 92.3 $\pm$ 4.5   & 95.0 $\pm$ 3.0 \\ 
PointCG~(Point-M2AE)       & \textbf{97.0} $\pm$ 1.9  & \textbf{98.4} $\pm$ 1.6  & \textbf{92.8} $\pm$ 3.8    & \textbf{95.5} $\pm$ 2.9 \\ 
\bottomrule
\end{tabular}
\label{tab:Cls_MN40_FewShot}
\end{table*}


\subsection{Part Segmentation} \label{sec:PartSeg}
\par The task of part segmentation aims to predict more fine-grained class labels for every instance. We conduct part segmentation on ShapeNetPart~\cite{yi2016scalable}, which comprises $16,881$ samples shared by $16$ categories, annotated with $50$ parts in total. 
As illustrated in Tab.~\ref{tab:partSegDetail}, our method achieves competitive results and outperforms others in eleven categories.
\begin{table*}[]
\caption{Part segmentation results on ShapeNetPart. We report the mean IoU across all part categories $mIoU_C(\%)$ and the mean IoU across all instances $mIoU_I(\%)$, as well as the IoU(\%) for categories.}
\centering
\scalebox{0.8}{
\begin{tabular}{c|cc|cccccccccccccccc}
\toprule
\multirow{2}{*}{Methods}  & \multirow{2}{*}{$mIoU_C$} & \multirow{2}{*}{$mIoU_I$}   & aero       & bag     & cap     & car     & chair     & ear     & guitar    & knife    & lamp     & laptop    & motor    & mug     & pistol    & rocket    & skate   & table \\  &       &       & plane     
 &        &         &         &           & phone   &            &         &          &          & bike     &         &           &           & board    &  \\ \hline
Transformer~\cite{yu2022pointBERT}     & 83.42   & 85.1    & 82.9     & 85.4      & 87.7     & 78.8           & 90.5  & 80.8      & 91.1       & 87.7            & 85.3      & 95.6   & 73.9  & 94.9  & 83.5   & 61.2   & 74.9    & 80.6  \\
Point-BERT~\cite{yu2022pointBERT}       & 84.11  & 85.6    & 84.3     & 84.8        & 88.0    & 79.8          & 91.0  & 81.7      & 91.6       & 87.9  & 85.2 & 95.6   & 75.6      & 94.7    & 84.3     & 63.4     & 76.3       & 81.5  \\
Point-MAE~\cite{pang2022masked}         & 84.19    & 86.1     & 84.3      & 85.0      & 88.3           & 80.5        & \textbf{91.3}   & 78.5      & \textbf{92.1}    & 87.4  & \textbf{86.1}    & \textbf{96.1}   & 75.2  & 94.6   & 84.7      & 63.5  & \textbf{77.1}    & \textbf{82.4}  \\
\textcolor{black}{PointCG~(Point-MAE)} & \textbf{84.48}  & \textbf{86.2}  & \textbf{84.4} & \textbf{86.1}  & \textbf{88.4}  & \textbf{80.7}  & \textbf{91.3}   & \textbf{81.2}   & 91.8   & \textbf{88.3}     & 85.9     & 95.9   & \textbf{75.7}   & \textbf{94.9}  & \textbf{85.1}    & \textbf{63.7}     & 76.5      & 81.8  \\ 
\bottomrule
\end{tabular}}
\label{tab:partSegDetail}
\end{table*}
\par
\begin{figure}[t]
    \centering
    \includegraphics[width=1.0\linewidth]{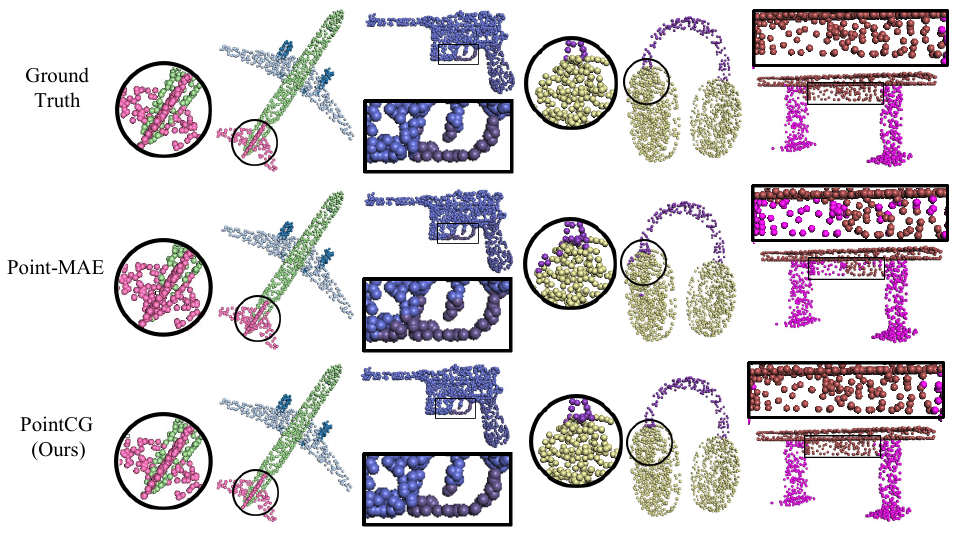}
    \caption{Visualization of the part segmentation results on the test set of ShapeNetPart. 
    }
	  \label{fig:multiClass_partSeg}
\end{figure}

 \textbf{Visualization of part segmentation}. Fine-grained part segmentation holds immense practical value. To highlight the clear advantage of our method in this task, we visualize the results and compare them with Point-MAE~\cite{pang2022masked} in Fig.~\ref{fig:multiClass_partSeg}. 
As depicted in the third line, our method accurately segments the fuselage and tail fin of the airplane, along with the earphone and headband. 
This reveals the capability of our method to capture discriminative features of points belonging to distinct sections within the same instance.

\subsection{Semantic Segmentation} \label{sec:SemSeg}
\par Large-scale indoor datasets introduce more complexities as they cover larger scenes in real-world environments with noise and outliers. 
We evaluate the performance of our pre-trained model on the 3D semantic segmentation task using the Stanford large-scale 3D Indoor Spaces (S3DIS)~\cite{armeni2016S3DIS} dataset. S3DIS includes data from $6$ indoor areas, comprising a total of $272$ rooms. 
We fine-tune the pre-trained model with Area 1-5 and evaluate it with Area 6. The results for each category are shown in Tab.~\ref{tab:semanticSeg_Class}. 
Our method outperforms Point-MAE~\cite{pang2022masked} across all categories except `beam' and `board'. The results underscore our model's capability to extract contextual and semantic information, which is crucial for producing fine-grained segmentation outcomes.

\begin{table*}[]
\caption{Semantic segmentation results on S3DIS. 
The mean class-wise intersection over union~($mIoU_C(\%)$), mean class-wise accuracy~($mAcc(\%)$), overall accuracy~($oAcc(\%)$), and the $IoU(\%)$ of each class are reported.
}
\centering
\scalebox{0.85}{
\begin{tabular}{c|ccc|ccccccccccccc}
\toprule
Methods                         & $oAcc$       & $mAcc$         & $mIoU_C$     & ceiling       & floor           & wall    & beam & column & window & door & table & chair & sofa  & bookcase  & board  & clutter  \\ \hline
Point-MAE~\cite{pang2022masked} & 90.29       & 82.18         & 73.14          & 94.4          & 96.8            & 77.6     & \textbf{82.0}  & 68.5          & 78.4            & 79.0       & 73.3             & 82.9          & 45.4             & 52.6      & \textbf{55.5}   & 64.6 \\
\textcolor{black}{PointCG~(Point-MAE)}  &\textbf{92.22} &\textbf{85.17} &\textbf{76.58}   & \textbf{96.5} & \textbf{98.4}  & \textbf{81.6}   & 81.9    & \textbf{82.0}  & \textbf{82.9}   & \textbf{83.5}  & \textbf{75.1} & \textbf{84.7}  & \textbf{49.6} & \textbf{57.5} & 52.6       & \textbf{69.1}  \\
\bottomrule
\end{tabular}}
\label{tab:semanticSeg_Class}
\end{table*}

In reference to the semantic segmentation experiments of STRL~\cite{huang2021STRL}, we fine-tune our pre-trained model on one area in Area 1-5, followed by evaluation on Area 6. We extend the experiments of Point-MAE~\cite{pang2022masked} based on the pre-trained model released in the official code and present the mean $IoU$ across all class categories $mIoU(\%)$ and the classification accuracy Acc (\%) in Tab.~\ref{tab:semSeg}. 
Our model exhibits a significant improvement in accuracy and $mIoU$ compared to STRL~\cite{huang2021STRL} and Point-MAE~\cite{pang2022masked}. These results demonstrate the capability of our model to extract contextual and semantic information, leading to fine-grained segmentation results.
\begin{table}[htbp]
\caption{Semantic segmentation on S3DIS.}
\centering
\begin{tabular}{cccc}
\toprule
Fine-tuning Area                      & Method       & Acc.    & mIoU  \\ \hline
\multirow{3}{*}{Area 1 (3687 samples)} & STRL~\cite{huang2021STRL}  & 85.28 & 59.15 \\  
& Point-MAE~\cite{pang2022masked}   & 89.03 & 71.92 \\  
& \textcolor{black}{PointCG~(Point-MAE)}  & \textbf{90.29} & \textbf{74.28}  \\ \hline
\multirow{3}{*}{Area 2 (4440 samples)}  & STRL~\cite{huang2021STRL} & 72.37 & 39.21 \\ 
& Point-MAE~\cite{pang2022masked}   & 76.72 & 47.13 \\   
& \textcolor{black}{PointCG~(Point-MAE)}  & \textbf{78.31} & \textbf{48.95}  \\ \hline
\multirow{3}{*}{Area 3 (1650 samples)}  & STRL~\cite{huang2021STRL} & 79.12 & 51.88 \\ 
& Point-MAE~\cite{pang2022masked}   & 84.09 & 64.29 \\   
& \textcolor{black}{PointCG~(Point-MAE)}  & \textbf{85.21} & \textbf{65.63} \\ \hline
\multirow{3}{*}{Area 4 (3662samples)}   & STRL~\cite{huang2021STRL} & 73.81 & 39.28 \\ 
& Point-MAE~\cite{pang2022masked}   & 77.34 & 45.15 \\  
& \textcolor{black}{PointCG~(Point-MAE)} & \textbf{78.07} & \textbf{47.30}\\ \hline
\multirow{3}{*}{Area 5 (6852 samples)}  & STRL~\cite{huang2021STRL} & 77.28 & 49.53 \\ 
 & Point-MAE~\cite{pang2022masked}   & 80.56 & 51.46 \\ 
& \textcolor{black}{PointCG~(Point-MAE)} & \textbf{81.79} & \textbf{54.04}\\ 
\bottomrule
\end{tabular}
\label{tab:semSeg}
\end{table}

\textcolor{black}{ 
\subsection{Indoor 3D object detection} \label{sec:ObjDetect}
To validate the effectiveness of our method in scene-level prediction tasks, we conduct object detection experiments on ScanNetV2~\cite{dai2017scannet}, a 3D indoor scene dataset with rich annotations, including 1,513 scenes across 18 object classes. The dataset includes semantic labels, per-point instances, and both 2D and 3D bounding boxes. 
Following TAP~\cite{Wang_2023_TAP}, we adopt 3DETR~\cite{3DETR_2021_ICCV} as a baseline, construct the PointCG network, and pre-train on the object-level dataset ShapeNet55~\cite{chang2015shapenet}. 
}
\par \textcolor{black}{
As shown in Tab.~\ref{tab:ObjDetect_Indoor}, our method demonstrates superior performance compared to both the baseline 3DETR~\cite{3DETR_2021_ICCV} and TAP~\cite{Wang_2023_TAP} in both $AP_{0.25}$ and $AP_{0.5}$ metrics. 
This improvement indicates that the encoder trained by PointCG effectively captures discriminative information and generalizes well to complex scenes even when pre-trained with object-level datasets.
}

\begin{table}[]
\caption{\textcolor{black}{Scene-level object detection on ScanNetV2~\cite{dai2017scannet}. 
Average precision at 0.25 IoU thresholds ($AP_{0.25}$) and 0.5 IoU thresholds ($AP_{0.5}$) of detection are reported.
}}
\centering
\begin{tabular}{lccc}
\toprule
Methods              & Pre-T  & $AP_{0.25}$  & $AP_{0.5}$ \\ \hline
VoteNet                 & - & 58.6  & 33.5   \\ 
3DETR                   & -  &  62.1 & 37.9    \\ 
3DETR+TAP        & ShapeNet  & 63.0(+0.9) & 41.4(+3.5) \\ 
3DETR+PointCG   & ShapeNet  & 63.21(+1.11) & 42.17(+4.27) \\
\bottomrule
\end{tabular}
\label{tab:ObjDetect_Indoor}
\end{table}

\subsection{Ablation Study} \label{sec:Ablation}
To investigate the architectural designs of our method, 
we conduct comprehensive ablation studies with Point-MAE as the backbone model and elucidate the individual contribution of each module. 

\par \textbf{Effectiveness of the components}. 
As shown in Tab.~\ref{tab:ablationModules}, we validate the effectiveness of each module by enhancing and replacing modules on the baseline.
We adopt Point-MAE~\cite{pang2022masked} for comparison and utilize hidden point completion~(HPC) as the baseline (a). 
In (b), we add the feature alignment module with pre-trained Vit-B/16~\cite{radford2021learning}.
Based on (b), we incorporate the arbitrary-view image generation~(AIG) module, constituting our PointCG, denoted as (c).
\textcolor{black}{As shown in Tab.~\ref{tab:ablationModules}, while the inclusion of the feature alignment module based on HPC does not substantially improve classification accuracy, the exclusion of this module from PointCG yields a diminished Linear-SVM accuracy of $88.41\%$.}
\begin{table}[]
\caption{Ablation studies on the introduced modules. Shape classification based on pre-trained models. 
}
\centering
\begin{tabular}{lccc}
\toprule
Methods                       & Linear-SVM.   & Acc.  & Acc+Vote \\ \hline
Point-MAE~(Rep.)~\cite{pang2022masked}  & - & 92.22 & 93.21    \\ 
(a) HPC   & 91.15 & 92.76 & 93.47    \\ 
(b) + Feature Alignment   & 91.23 & 92.99 & 93.51    \\ 
(c) + AIG   & 92.26 & 93.52 & 94.03   \\ 
(d) PointCG~(Vit-B/32)   & 92.21 & 93.05 & 93.97   \\ 
(e) PointCG~(ResNet50~\cite{he2016deep})  & 91.01 & 92.75 & 93.39 \\ 
(f) PointCG~(Grayscale image)  & 91.75 & 93.19 & 93.52   \\ 
(g) PointCG~(depth map)       & 91.09 & 92.78 & 93.43   \\ 
\bottomrule
\end{tabular}
\label{tab:ablationModules}
\end{table}
We further replace Vit-B/16 with Vit-B/32~(d) and ResNet50~\cite{he2016deep}~(e). While Vit-B/32 outperforms Vit-B/16 in the image domain, it does not improve the accuracy of classification. The model with ResNet50~\cite{he2016deep} exhibits comparatively poorer performance. 

To assess the impact of color in rendered images, we extract grayscale images from the rendered ones and pre-train with them in (f). This operation leads to a decrease in shape classification. 
Grayscale images may potentially lose structural or finer details inherent in the original images.
In experiment~(g), we extract depth maps from point clouds following PointCLIP~\cite{zhang2022pointclip} and pre-train with them instead of rendered images. The classification result shows poor performance. 

\par \textcolor{black}{\textbf{Training and inference time.} Tab.~\ref{tab:Ablation_pretrain_inference_time} presents pre-training and inference times for the classification task of each module. The results demonstrate that the pre-training time is notably longer than that of the baseline, whereas the inference time for classification remains comparable.}
\par \textcolor{black}{The pre-training of HPC requires approximately $397s$ per epoch. 
Compared to the baseline~(Point-MAE), the primary time consumption arises from the data organization module, which is responsible for obtaining visible points from arbitrary views as inputs. 
Among the components, the 3D completion module has the shortest pre-training duration, while the feature alignment and AIG modules demand considerably more time.
}

\begin{table}[]
\caption{\textcolor{black}{Ablation studies on the introduced modules. Training and inference times of each model. }
}
\centering
\scalebox{0.95}{
\begin{tabular}{lcc}
\toprule
Methods                  & Epoch time~(s)  & Inference time~(s) \\ \hline
Point-MAE~(Rep.)~\cite{pang2022masked} & 54.73-57.59  & 36.85   \\ 
    HPC                            & 395.18-399.26    & 36.65  \\ 
PointCG W/O AIG                    & 413.96-419.47    & 37.65  \\
PointCG W/O Feature Alignment      & 405.52-411.22    & 35.67  \\
PointCG W/O 3D Completion          & 520.97-524.73    & 36.79  \\
PointCG~(Point-MAE)                & 534.56-538.72    & 37.37 \\ 
\bottomrule
\end{tabular}}
\label{tab:Ablation_pretrain_inference_time}
\end{table}

\begin{figure}[htbp]
    \centering
    \includegraphics[width=1.0\linewidth]{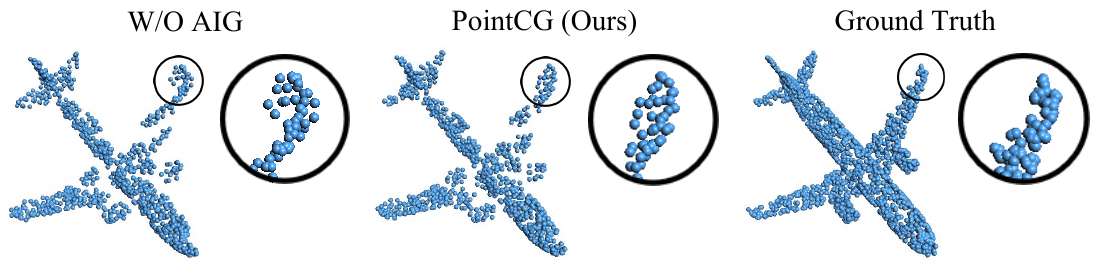}
    \caption{ 
    Visualization of the 3D completion results based on the pre-trained models of PointCG without AIG~(W/O AIG) and PointCG on ModelNet40~\cite{wu2015_modelNet}. 
}
    \label{fig:Rec_MN40_WO_ImgGen}
\end{figure}

\par \textbf{Visualization of 3D Completion.}
To validate the positive impact of AIG on the 3D completion task, we exclude AIG from PointCG~(W/O AIG) and pre-train the network. The 3D completion results of this variant and PointCG are shown in Fig.~\ref{fig:Rec_MN40_WO_ImgGen}.
As depicted, the edges of the aircraft wings are sharpened with our method. Without AIG, the completion edges exhibit point groups, attributed to solely relying on 3D completion, while the completion targets are point clusters. 

\begin{figure}[htbp]
    \centering
    \includegraphics[width=1.0\linewidth]{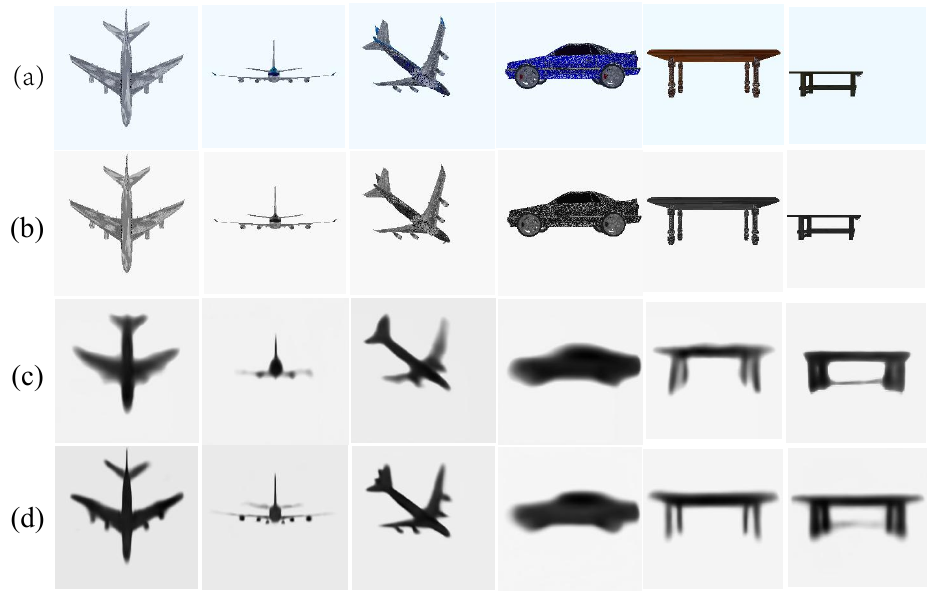}
    \caption{Visualization of the generated images based on pre-trained models. 
     We present the original rendered images in row (a) and corresponding grayscale one in row (b). The row (c) shows the results of PointCG without 3D completion. The results based on PointCG are shown in row (d).}
    \label{fig:Ablation_GenImgs_WO_3DRec}
\end{figure}
\par \textbf{Visualization of the generated images.} 
To validate the provision of geometric structural information by 3D completion, we showcase the results of image generation after excluding the completion module and compare them with PointCG's results in Fig.~\ref{fig:Ablation_GenImgs_WO_3DRec}.
As depicted in line (c), outcomes exhibit significant artifacts and lack local structural information. Specifically, in the case of the aircraft tail wing, only the wing's shape is generated, omitting volumetric structural details. However, our method in line (d) successfully predicts the accurate shapes of the objects.

\par \textbf{Camera perspectives}. 
To identify more suitable inputs and predicted images, we design experiments with different camera positions as inputs, and the results are reported in Tab.~\ref{tab:ablation_cameraViews}. 
(a) takes the left-view image as input and the front-view as target, and (b) takes the left-view image as input and an arbitrary-view image as target. 
In (c), we use images from three different views (front, left, and top views) as inputs and an arbitrary-view image as target. 
Both the input and target of (d) are from arbitrary views, yielding the optimal results. 
Therefore, we utilize two arbitrary-view images as the input and target, respectively. 
\begin{table}[]
\caption{\textcolor{black}{Ablation studies on camera views. Pre-training and shape classification with images from different views.}}
\centering
\scalebox{0.9}{
\begin{tabular}{lccc}
\toprule
Input/Prediction                             & Linear-SVM.  & Acc.   & Acc+Vote \\ \hline
(a) Left view /front view                       & 91.13       & 92.66  & 93.31    \\ 
(b) Left view /arbitrary view                   & 91.65       & 92.83  & 93.68    \\ 
(c) Front, left, and top views/arbitrary view   & 92.22       & 93.03  & 93.40   \\ 
(d) Arbitrary view/arbitrary view               & 92.26       & 93.52  & 94.03   \\ 
\bottomrule
\end{tabular}}
\label{tab:ablation_cameraViews}
\end{table}

\par \textbf{Pre-training with more complete inputs}. 
We posit that if the inputs of 3D completion contain more structural information and more overlap areas with the targets, the completion task will be accomplished more easily. This leads to a reduced training intensity for the backbone, thereby diminishing the backbone's perception of 3D objects.
We design this study based on different inputs. 
The inputs in (a) are derived from a single arbitrary view. The inputs for (b) and (c) involve the addition of two and eight extra patches, respectively, to the input of (a). The points from two arbitrary views serve as the inputs for (d). Results are reported in Tab.~\ref{tab:Ablation_Inputs}. 

\begin{table}[h]
\caption{Ablation study on the completeness of inputs. We pre-train the model based on inputs with varying levels of completeness.
}
\centering
\scalebox{0.9}{
\begin{tabular}{lccc}
\toprule
  Input                         & Linear-SVM.  & Acc.   & Acc+Vote \\ \hline
(a) One arbitrary view          & 92.26    & 93.52  & 94.03   \\ 
(b) One arbitrary view + 2 patches  & 91.73   & 93.14  & 93.55    \\ 
(c) One arbitrary view + 8 patches & 91.41   & 92.85  & 93.25   \\ 
(d) Two arbitrary views                & 91.02   & 92.75  & 93.27   \\ 
\bottomrule
\end{tabular}}
\label{tab:Ablation_Inputs}
\end{table}

In cases (b) and (c), the accuracy of the Linear-SVM during pre-training decreases as more patches are included in the inputs. 
The classification results via fine-tuning also show a decline. 
In case (d), inputs from two views offer more structural information about the input objects, which leads to a significant decrease in shape classification. 
\textcolor{black}{
This experiment reveals that as additional structural information is progressively included in the input, shape classification accuracy steadily decreases. This indicates that excessive exposure to object structure within the inputs hinders the model's learning ability. }

\par
\begin{figure}[htbp]
    \centering
    \includegraphics[width=1.0\linewidth]{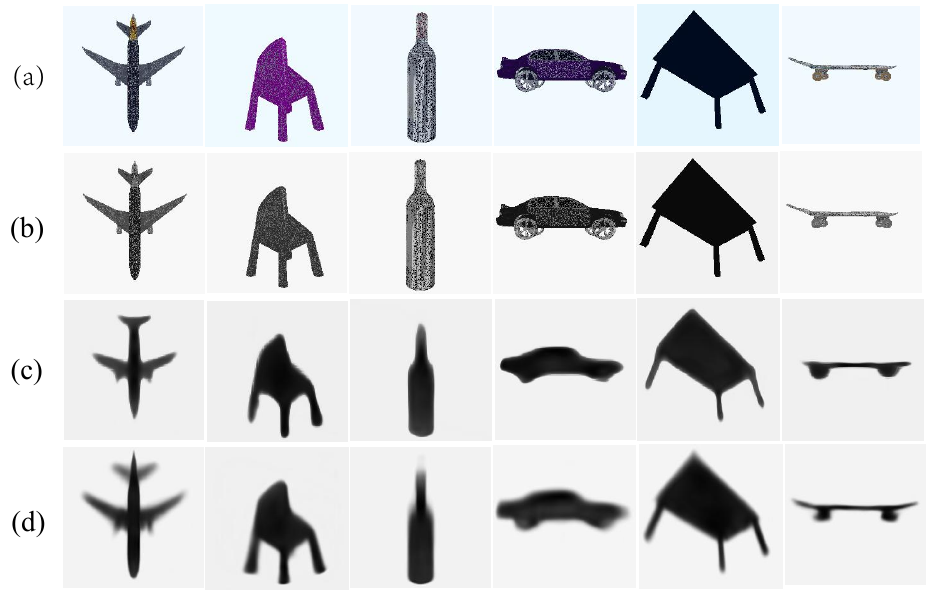}
    \caption{Visualization of the generated images with different losses. 
     We present the original rendered images in row (a) and corresponding grayscale versions in row (b). Rows (c) and (d) are generated with the mixed losses of $1.0*\mathcal{L}_1+1.0*(1.0-\mathcal{L}_{MS\_SSIM})$ and $1.0*\mathcal{L}_1+0.2*\mathcal{L}_{MSFR}$, respectively.
    }
    \label{fig:Ablation_GenImgs_losses}
\end{figure}

 \textbf{Image generation losses}.  
AIG significantly impacts the backbone's perception of 3D objects by precise supervision between the ground truth and the generated images. It always affects the quality of the generated images~(as shown in Fig.~\ref{fig:Ablation_GenImgs_losses}).
We conduct experiments to examine the performance of various loss functions for supervision, as outlined in Tab.~\ref{tab:ablation_GenImgLoss}.  
\begin{table}[]
\caption{Ablation study on the loss functions of image generation. We pre-train with various loss functions and subsequently fine-tune with shape classification.
}
\centering
\begin{tabular}{lccc}
\toprule
Generation Loss                         & Linear-SVM   & Acc.   & Acc+Vote\\ \hline
$\mathcal{L}_1$                         & 90.32        & 93.07   & 93.43    \\ 
$\mathcal{L}_2$                         & 91.13        & 92.97   & 93.40    \\ 
$1.0*\mathcal{L}_1+1.0*\mathcal{L}_2$    & 91.97       & 93.35   & 93.76   \\ 
$\mathcal{L}_1+(1.0-\mathcal{L}_{MS\_SSIM})$           & 91.73       & 93.23   & 93.61  \\ 
$0.8*\mathcal{L}_1+0.2*\mathcal{L}_{MSFR}$             & 91.65       & 93.48   & 93.81 \\ 
$1.0*\mathcal{L}_1+0.2*\mathcal{L}_{MSFR}$             & 92.26       & 93.52   & 94.03  \\ 
\bottomrule
\end{tabular}
\label{tab:ablation_GenImgLoss}
\end{table}

\par The $\mathcal{L}_2$ loss penalizes large errors more heavily and is more tolerant of small errors. 
In contrast, the $\mathcal{L}_1$ loss does not excessively penalize large errors. 
The classification results of the model with the $\mathcal{L}_1$ loss yield superior results compared to the $\mathcal{L}_2$ loss. 

\par The multi-scale structural similarity~(MS\_SSIM) index preserves the contrast in high-frequency regions. While $\mathcal{L}_1$ is effective in preserving colors and luminance~\cite{Zhao_2017_Losses}, but does not produce quite the same contrast as MS\_SSIM. To leverage the benefits of both loss functions, we utilize the combination of them: $\mathcal{L}_{G}=\alpha*{\mathcal{L}_1}+\beta*\mathcal{L}_{MS\_SSIM}$. However, the classification quality is slightly lower than others. 
As shown in this table, the model with $\mathcal{L}_{G}=1.0*\mathcal{L}_1+0.2*\mathcal{L}_{MSFR}$ achieves the best classification results
\par \textbf{Visualization of the generated images.} To assess a more suitable image generation loss for our model, we visualize the generated images with different losses in Fig.~\ref{fig:Ablation_GenImgs_losses}.
Despite minor artifacts near the edges or corners, our model consistently produces complete and structurally clear images in line (d). 
This indicates that our model possesses the capability to capture geometric structures and stereoscopic knowledge about 3D objects, as well as the ability to infer the occluded points from arbitrary views.


\textcolor{black}{\textbf{Qualitative results of 2D image generation. } 
While Figs.~\ref{fig:Ablation_GenImgs_WO_3DRec} and \ref{fig:Ablation_GenImgs_losses} provide visual examples of the generated images, they lack rigorous qualitative results to substantiate any significant improvements over baseline methods in terms of preserving 3D structure in 2D images. To address this, we present Tab.~\ref{tab:Ablation_GenImgs_Quant}, which provides quantitative evaluations of the generated images under various architectures and image generation losses. 
}
\par \textcolor{black}{
We employ Mean Squared Error~(MSE), Structural Similarity index~(SSIM), Peak Signal to Noise Ratio~(PSNR), and Normalized Mutual Information~(NMI) as our primary image evaluation metrics.
Clearly, the configuration of PointCG with $1.0*\mathcal{L}_1+0.2*\mathcal{L}_{MSFR}$~(d) achieves the best performance across all metrics. By contrast, PointCG without 3D Completion~(a) exhibits the poorest results. This indicates that the 3D completion module significantly enhances the quality of the generated images.}

\begin{table}[]
\caption{\textcolor{black}{Qualitative results of the generated images based on different architectures and image generation losses.
}}
\centering
\scalebox{0.9}{
\begin{tabular}{lcccc}
\toprule
Methods/Generation Loss               & MSE~$\downarrow$  & PSNR~$\uparrow$   & SSIM~$\uparrow$  & NMI~$\uparrow$  \\ \hline
(a) PointCG W/O 3D Completion     & 0.054  & 30.099  & 0.795 & 0.485 \\ 
(b) $1.0*\mathcal{L}_1+1.0*\mathcal{L}_2$ & 0.038  & 31.599  & 0.831 & 0.515 \\ 
(c) $\mathcal{L}_1+(1.0-\mathcal{L}_{MS\_SSIM})$  & 0.042   & 31.519  & 0.827   & 0.509  \\ 
(d) $1.0*\mathcal{L}_1+0.2*\mathcal{L}_{MSFR}$   & \textbf{0.034}  & \textbf{32.106} & \textbf{0.856}    & \textbf{0.558}  \\ 
\bottomrule
\end{tabular}}
\label{tab:Ablation_GenImgs_Quant}
\end{table}


\section{Conclusion} \label{sec:Conculsion}
\par In this paper, we propose PointCG, a unified framework with hidden points completion and arbitrary-view image generation for self-supervised point cloud learning. 
Completion and generation based on partial points prompt the encoder to extract high-quality representations with 3D structural intricacies and alleviate ambiguous supervision. 
Thereby, our method achieves notable enhancements over baseline methods and outperforms similar methods in classification and reconstruction tasks on real datasets.
We expect our pre-trained models will benefit a wide range of 3D tasks, including 3D object detection, semantic segmentation, and visual grounding.


While performing well across multiple tasks, there is much room for improving PointCG. This includes expanding the modality of images to incorporate other formats such as language or audio. 
Furthermore, while we focus on instance-level tasks, scene-level understanding is crucial in real-world applications. 
Therefore, our future studies may involve delving into applications in scene understanding and investigating interactions among multiple modalities for 3D reasoning. 


\bibliographystyle{IEEEtran}
\bibliography{IEEEfull}
 
\vspace{11pt}

\vspace{11pt}

\vfill

\end{document}